\documentclass[letterpaper]{article}
\usepackage{aaai2027}
\usepackage[hyphens]{url}
\usepackage{graphicx}
\usepackage{natbib}
\usepackage{caption}
\usepackage{booktabs}
\usepackage{amsmath}
\usepackage{amssymb}

\title{Think Before You Transcribe: Context-Steered Reasoning for Hotword-Aware Speech Recognition}
\author{
    Qihao Wang\textsuperscript{\rm 1,\rm 2},
    Hang Su\textsuperscript{\rm 3},
    Fan Cui\textsuperscript{\rm 3},
    Lichun Fan\textsuperscript{\rm 3},
    Jian Luan\textsuperscript{\rm 3},
    Yue Hu\textsuperscript{\rm 1,\rm 2}
}
\affiliations{
    \textsuperscript{\rm 1}Institute of Information Engineering, Chinese Academy of Sciences\\
    \textsuperscript{\rm 2}School of Cyber Security, University of Chinese Academy of Sciences\\
    \textsuperscript{\rm 3}MiLM Plus, Xiaomi Inc.\\
}
\newcommand{\CAST}{\textsc{CAST}}
\newcommand{\RACE}{\textsc{RACE}-Bench}
\newcommand{\trp}[3]{#1/#2/#3}
\newcommand{\best}[1]{\textbf{#1}}
\newcommand{\second}[1]{\underline{#1}}
\newcommand{\trpbest}[3]{\best{#1}/\best{#2}/\best{#3}}

\newcommand{\metriclegend}{WER / NE-WER / NE-FNR (\%; $\downarrow$)}

\begin{document}
\maketitle

\begin{abstract}
Contextual automatic speech recognition (ASR) improves rare-entity recognition by supplying candidate hotwords, but retrieved candidates may be incomplete or incorrect. A reliable system must therefore exploit a relevant candidate without allowing unsupported alternatives to override the acoustic evidence. We propose a context-steered reasoning framework that generates a short, audio-grounded caption before transcription. This task-specific representation lets the model assess candidate relevance before making a lexical decision. Supervised fine-tuning establishes the structured caption-to-transcription behavior, while targeted trajectory injection strengthens group relative policy optimization when difficult utterances yield only unsuccessful rollouts. Experiments across diverse context conditions show that the proposed method consistently improves both transcription quality and hotword recognition while remaining robust to irrelevant candidates.
\end{abstract}

\section{Introduction}

Modern ASR systems, such as Whisper \cite{radford2023whisper}, achieve high accuracy on general-domain speech, yet remain unreliable on rare names and specialized terminology in medical, legal, financial, and technical applications \cite{SeedASR}. These expressions are sparsely represented in general training data, often acoustically confusable, and disproportionately important to the utterance. Because such errors can invalidate otherwise readable transcripts and downstream search, rare-entity recognition remains a persistent obstacle to deploying ASR in knowledge-intensive settings.

Contextual ASR mitigates this problem by supplying candidate hotwords as an additional prior. Conventional systems integrate this information through shallow fusion, contextual adapters, specialized biasing modules, or external memories \cite{yu2024lcbnetlongcontextbiasingaudiovisual,SpeechMemory}; recent audio-language models instead place candidate entities directly in the prompt \cite{SeedASR,contextasr_with_Retrieval,ContextualASR_via_Fusion,qwen3omni}. Both paradigms improve rare-entity recognition when the supplied context is accurate. In deployment, however, hotwords are retrieved from a large inventory rather than curated for each utterance. Deeper retrieval improves target coverage but also introduces more plausible distractors, so the list may omit a spoken entity or contain no relevant entity at all. Context is thus uncertain evidence rather than a trusted constraint.

This retrieval setting changes the problem from unconditional biasing to selective use of uncertain evidence. A model that discounts the list can still miss a relevant candidate, whereas a model that trusts it too readily can insert an unsupported entity. Yet most contextual ASR benchmarks emphasize curated ground-truth lists or a single domain. A controlled evaluation should instead vary candidate quality for the same utterance, separating gains from relevant context from degradation caused by unsupported context. Without such pairing, it is unclear whether an improvement reflects better speech recognition, stronger contextual integration, or merely an easier candidate list.

The same uncertainty exposes a limitation in direct transcription. Mapping audio and candidates immediately to the final text leaves no explicit stage at which the model can relate a plausible lexical form to the acoustic and semantic content before committing to it. Weak steering may miss an ambiguous entity, while excessive steering copies a plausible but unsupported candidate. Intermediate reasoning has proved useful in language and audio tasks \cite{deepseekr1,audioCoT,SoundMind}; for contextual ASR, it can instead serve the focused role of grounding supported candidates in a description of the utterance while leaving unrelated candidates unused.

Learning this selective behavior also creates an optimization problem. Relative policy objectives are informative when a rollout group contains outputs of different quality, but every rollout for a difficult utterance may miss a supplied target entity. GRPO can rank these failures, yet the group contains no trajectory that demonstrates how a successful caption should connect the audio, candidate, and final lexical choice.

We propose \textbf{CAST} (Contextual ASR with Steered Thinking), a framework that produces a natural-language caption before the transcript. The caption summarizes the perceived utterance and naturally mentions only candidates that the model finds supported before it commits to a lexical choice. We train this behavior with supervised fine-tuning (SFT) followed by GRPO \cite{deepseekmath}. During GRPO, targeted off-policy trajectory injection handles eligible all-failure groups by probabilistically replacing the lowest-reward rollout with a reference-conditioned caption-to-transcription trajectory generated by the SFT checkpoint. This replacement supplies a concrete positive example while retaining the remaining sampled failures for relative comparison. The resulting mixed group is processed by the usual group-normalized loss. Because the injected item is not sampled under the original policy input and receives no importance correction, the update is a deliberately biased training heuristic rather than an on-policy GRPO estimate.

We introduce \textbf{RACE-Bench} (Robust ASR with Contextual Entities Benchmark) to evaluate the same model as context quality changes. The benchmark contains 9,136 real and synthetic English and Chinese utterances, all reserved for testing. Each utterance is paired with correct-only, contextless, retrieved, and all-noise candidate lists. This paired design measures not only whether a model can exploit a supplied entity, but also how its behavior changes as useful context disappears and distractors remain. Because the audio is fixed across conditions, the results separate effective use of available entities from resistance to unsupported candidates.

On \RACE, caption-based reasoning outperforms direct-transcription variants trained on the same data, and targeted injection further improves an otherwise identical GRPO policy. \CAST\ gains from larger retrieved candidate sets without becoming more susceptible to all-noise context. Comparisons with native reasoning models show that the benefit comes from the task-specific intermediate representation rather than reasoning length alone. Its performance also transfers to ContextASR-Bench, while its output remains much shorter than unrestricted audio reasoning.

The main contributions of this work are:
\begin{enumerate}
    \item We propose \CAST, a caption-then-transcription framework that explicitly grounds candidate entities in the perceived audio content before the final lexical decision.
    \item We introduce targeted off-policy trajectory injection, supplying a positive trajectory when every GRPO rollout misses a target entity.
    \item We construct \RACE, a 9,136-utterance bilingual benchmark that provides a controlled evaluation of contextual ASR under correct-only, contextless, retrieved, and all-noise context on the same audio.
\end{enumerate}

\section{Related Work}

\noindent\textbf{Contextual ASR.}
Contextual ASR incorporates hotword lists, domain information, conversation history, or external memories to improve recognition of rare and domain-specific entities. Traditional systems use shallow fusion, contextual adapters, cross-attention, or explicit entity copying \cite{SeedASR,yu2024lcbnetlongcontextbiasingaudiovisual,OWSMBiasing,SpeechMemory,CopyNE,ContextualASR_via_Fusion}. Audio-language models can instead accept candidate entities directly in the prompt \cite{qwen2audio,qwen25omni,qwen3omni}. Retrieval-based approaches obtain relevant documents, speech exemplars, or acoustically similar entities before decoding \cite{LewisRAG2020,DOCRAG,LARAG,PhoneticRetrievalASR,contextasr_with_Retrieval,GLCLAP}, while LLM-based correction systems revise first-pass transcripts or filter retrieved candidates \cite{HyPoradise,ClozeGER,DeRAGEC}. Contextual ASR resources include SlideSpeech, ContextASR-Bench, and Contextual Earnings-22 \cite{slidespeech,ContextASR-Bench,ContextualEarnings22}.

\noindent\textbf{Audio Reasoning Models.}
Chain-of-thought reasoning generates intermediate steps before a final answer and has been studied in general language models \cite{wei2022chain,o1,deepseekr1} and multimodal reasoning \cite{visualcot,llavaCot}. Large audio and omni language models align audio encoders with language-model backbones, enabling end-to-end speech recognition and audio understanding \cite{qwen2audio,MiDashengLM,qwen25omni,qwen3omni,minicpmo,Megrez,baichuan-omni-1.5}. Recent work extends intermediate reasoning to audio understanding \cite{audioCoT,SoundMind,MindWithEyes} and speech recognition, including contextual and target-speaker ASR \cite{SpeechLLMsContextual,CocktailCoT}.

\begin{figure*}[t]
\centering
\includegraphics[width=\textwidth]{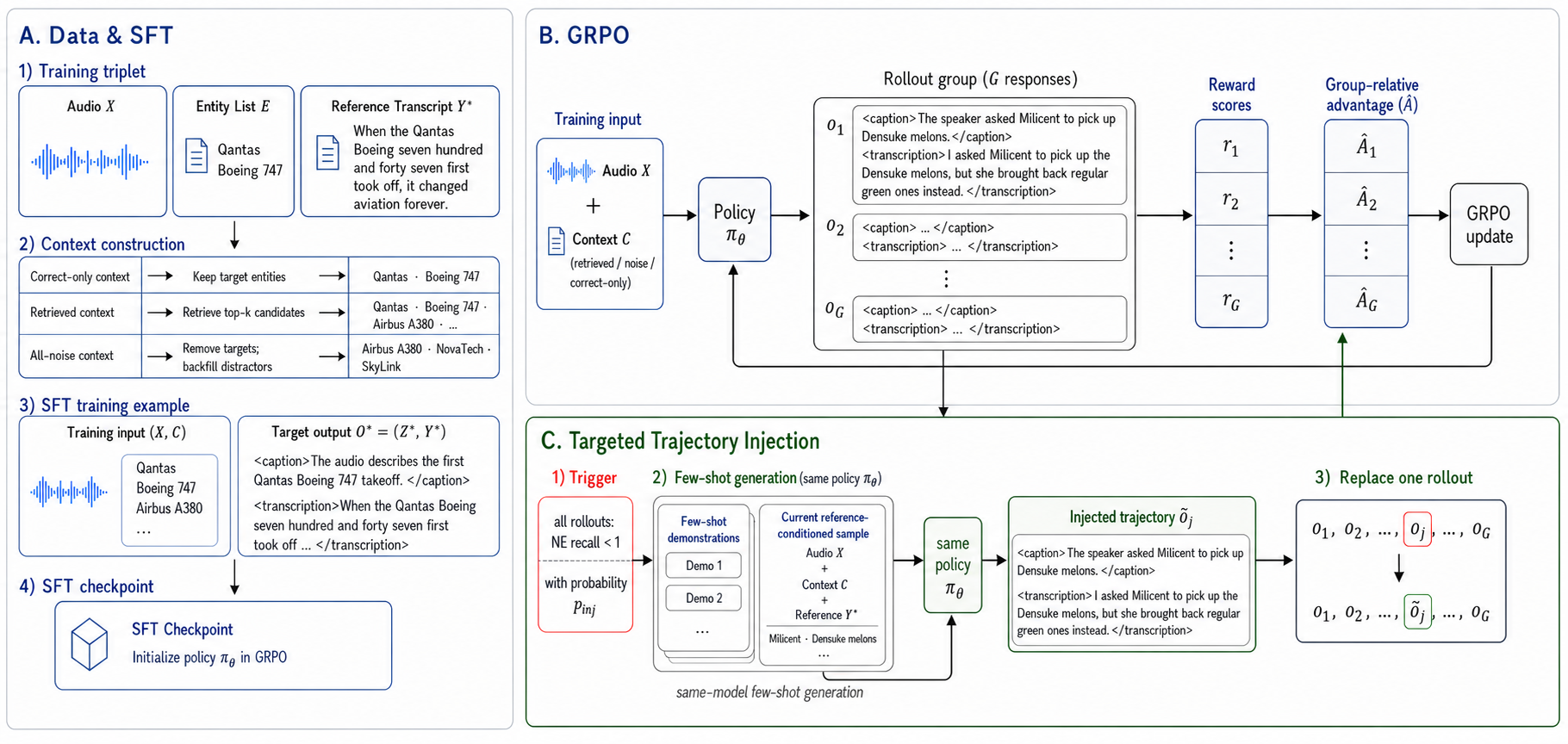}
\caption{Overview of the \CAST\ framework.}
\label{fig:overview}
\end{figure*}

\section{\RACE}
\label{sec:race}

\RACE\ contains 9,136 English and Chinese audio utterances spanning real and synthetic speech. It is designed to evaluate contextual ASR as candidate quality changes while the underlying audio remains fixed.

\subsection{Multi-Stage Data Curation}

For real speech, an LLM first extracts only explicit names, organizations, products, technical terms, and other proper entities from each reference, producing an $\langle$audio, transcript, entities$\rangle$ triplet. Qwen3-ASR-1.7B \cite{qwen3asr} then transcribes the audio without context. We retain samples with a nonzero named-entity false-negative rate (NE-FNR), more than ten transcript tokens, and a word error rate (WER) below 20\%. These filters select audio that is intelligible at the utterance level but remains difficult at the entity level.

Quality control then uses Qwen3-Omni-30B-A3B-Instruct (abbreviated below as Qwen3-Omni-30B-Instruct) with correct-only context. An LLM judge reviews disagreements between the contextual output and the annotation, distinguishing likely annotation errors from normalization, disfluency, repetition, and pronunciation variation. Proposed corrections update only the disputed entity and its occurrence in the transcript while preserving the original fields for traceability. A 50-sample audit found approximately 90\% of proposed corrections accurate; the main failure mode was replacing an audio-faithful pronunciation with a canonical spelling.
% TODO(author): add the full manual-review category distribution when provided.

Synthetic speech follows four controlled stages. We first generate natural spoken sentences containing one to three difficult entities, including uncommon names, transliterations, alphanumeric models, and domain terms. We normalize numbers, dates, symbols, and abbreviations to match their spoken realizations, then require every annotated entity to be an exact substring of the normalized reference. A multi-speaker TTS model generates the audio. Finally, contextual and contextless ASR checks remove synthesis failures and retain samples that are acoustically valid but challenging at the entity level. This pipeline provides controlled English and Chinese hard cases while maintaining agreement between the transcript and audio.

\subsection{Context Construction}

For each triplet with entity set $E$, we derive four views: \emph{correct-only} uses $E$; \emph{contextless} uses an empty list; \emph{RAG} retrieves top-$k$ candidates for $k\in\{5,10,20,50\}$; and \emph{all-noise} removes every retrieved member of $E$ and backfills with the next-ranked candidates to preserve list length. Every utterance is evaluated in every condition. RAG averages are arithmetic means over the four values of $k$.

We use GLCLAP \cite{GLCLAP} only as a fixed audio-to-text retrieval tool. Retrieval is closed-set over the deduplicated entity pool for the corresponding source and language. Its entity recall is 76.8/83.6/88.6/92.5\% for Chinese and 69.8/79.6/86.7/92.4\% for English at $k=5/10/20/50$, respectively. These imperfect recalls create realistic partial-context conditions, while rank-preserving backfill makes all-noise lists acoustically plausible rather than randomly unrelated.

Complete construction details, including subset composition, filtering and correction procedures, synthetic-speech generation, source-level entity-pool construction, and all-noise backfilling, are provided in the appendix.

\subsection{Comparison with Existing Benchmarks}

Existing resources provide multilingual, domain-specific, or naturally contextualized speech, but they do not pair correct-only, retrieved, and entirely incorrect candidate lists for the same utterances. As summarized in Table~\ref{tab:benchmark_comparison}, \RACE\ complements these benchmarks with a controlled comparison across candidate quality while holding the audio constant.

\begin{center}
\centering
\small
\setlength{\tabcolsep}{1.8pt}
\begin{tabular}{@{}lccccc@{}}
\toprule
Benchmark & Audio & Multi. & Oracle & Ret. & Noise \\
\midrule
CtxASR-B. & TTS & \checkmark & \checkmark & -- & -- \\
Earnings-22 & Real & -- & -- & -- & -- \\
SlideSpeech & Real & -- & -- & -- & -- \\
\RACE\ (ours) & Mixed & \checkmark & \checkmark & \checkmark & \checkmark \\
\bottomrule
\end{tabular}
\captionof{table}{Coverage of contextual ASR benchmarks: ContextASR-Bench \cite{ContextASR-Bench}, Contextual Earnings-22 \cite{ContextualEarnings22}, SlideSpeech \cite{slidespeech}, and \RACE. ``Multi.'' denotes multilingual coverage; ``Oracle'' denotes a correct-only entity list, ``Ret.'' retrieval-augmented lists, and ``Noise'' entirely incorrect lists.}
\label{tab:benchmark_comparison}
\end{center}

\FloatBarrier

\section{The \CAST\ Framework}
\label{sec:method}

\CAST\ has two components. A caption before transcription records how the model relates the candidate list to the audio. Targeted trajectory injection supplies a successful example when ordinary GRPO sampling produces no entity-complete rollout.

\subsection{Contextual ASR with Retrieved Candidates}

In retrieval-based contextual ASR, a system receives a speech signal $X$ together with a candidate list $C=\{c_1,\ldots,c_N\}$ and produces a transcription $Y$. Unlike a curated hotword list, $C$ is only a noisy prior: it may contain some, all, or none of the entities spoken in $X$. The system must therefore preserve acoustic fidelity when the list is unhelpful while exploiting a candidate when it resolves genuine acoustic ambiguity. \CAST\ is designed for this selective integration problem. At inference time it observes only $X$ and $C$; references and entity annotations are confined to training objectives and offline trajectory construction.

\subsection{Context-Grounded Reasoning}

Direct contextual ASR estimates $p_\theta(Y\mid X,C)$ and commits to each lexical decision while jointly interpreting the audio and candidate list. This implicit steering can be too weak to recover an acoustically ambiguous entity or too strong when an irrelevant candidate appears plausible. \CAST\ introduces a caption $Z$ before the transcription:
\begin{equation}
p_\theta(Z,Y\mid X,C)
=p_\theta(Z\mid X,C)\,
p_\theta(Y\mid X,C,Z).
\end{equation}
The caption is a natural-language description of the audio that mentions only candidates supported by the reference during target construction; supported entities are marked in the sentence rather than emitted as a separate selection list. The complete response has the form
\begin{equation}
\begin{split}
&\texttt{<caption>}Z\texttt{</caption>}\\[-2pt]
&\texttt{<transcription>}Y\texttt{</transcription>}.
\end{split}
\end{equation}
The model therefore grounds useful candidates in an acoustic interpretation before producing the transcription. This intermediate step strengthens the influence of supported entities without requiring every retrieved candidate to determine the output. The caption receives no separate reward: its value is learned through the quality and entity coverage of the resulting transcription. The appendix details training-corpus construction, caption-target generation, model instructions, and output parsing.

\subsection{Supervised Fine-Tuning}

We initialize \CAST\ from Qwen3-Omni-30B-Instruct \cite{qwen3omni}. Supervised fine-tuning provides a cold start for the two-part output format. If $O^\star=(Z^\star,Y^\star)$ is the caption-transcription target, the objective is standard next-token prediction:
\begin{equation}
\mathcal{L}_{\mathrm{SFT}}
=-\sum_{t=1}^{|O^\star|}
\log p_\theta(o^\star_t\mid X,C,o^\star_{<t}).
\end{equation}
This stage teaches the model to produce a caption before the transcript, but does not directly optimize WER or entity recall.

\subsection{Group Relative Policy Optimization}

Starting from the SFT checkpoint, we apply GRPO \cite{deepseekmath,GRPOSpeech,ContextualBiasingRL}. For each input $(X,C)$, the policy samples a group of outputs $O_i=(Z_i,Y_i)$. Only the transcription $Y_i$ contributes to the reward:
\begin{equation}
R_i=\lambda_{\mathrm{word}}R_{\mathrm{word}}(Y_i,Y^\star)
+\lambda_{\mathrm{ent}}R_{\mathrm{ent}}(Y_i,E^+),
\label{eq:reward}
\end{equation}
where $R_{\mathrm{word}}$ is clipped transcription accuracy derived from WER and $R_{\mathrm{ent}}$ is normalization-aware recall over the ground-truth entities that are present in the candidate list, denoted by $E^+$. The word-level term preserves overall fidelity, while the entity term increases the learning signal on acoustically ambiguous targets. No separate format, caption, length, or hallucination reward is used.\footnote{All experiments use $\lambda_{\mathrm{word}}=0.8$, $\lambda_{\mathrm{ent}}=0.2$, and a rollout group size of $G=8$.} The appendix gives the exact normalization, multiplicity, empty-set, and malformed-output rules.

Following standard GRPO, we normalize the total rewards within each group to obtain token-level advantages, optimize the clipped policy-ratio objective, and regularize against the SFT reference policy with a Kullback--Leibler (KL) penalty. Implementation hyperparameters are reported in Experimental Setup.

\subsection{Targeted Off-Policy Trajectory Injection}
\label{sec:inject}

Standard GRPO assumes that relative differences within a sampled group provide a useful learning signal. Contextual ASR exposes a specific failure of this assumption: for a difficult input, every rollout may miss at least one supplied target entity. Although the failures can still be ranked, none demonstrates how the entity should be grounded in the caption and carried into a correct transcription. Targeted injection closes this optimization gap by placing one successful trajectory into precisely such a group.

For rollout $i$, let $q_i$ denote recall over the provided target entities $E^+$. We trigger injection when $E^+$ is nonempty and
\begin{equation}
\max_{i\in\{1,\ldots,G\}}q_i<1.
\label{eq:trigger}
\end{equation}
Thus, even the best rollout in the group is incomplete. For an eligible group, we inject with probability $p=0.5$; otherwise the group follows ordinary GRPO. Contexts without a provided target entity do not satisfy the eligibility condition.

After SFT and before GRPO, we construct one stored trajectory for each training example using the SFT checkpoint $\pi_{\mathrm{SFT}}$. A three-shot prompt demonstrates how an audio signal, candidate list, and reference transcript map to the required caption. For the current example, $\pi_{\mathrm{SFT}}$ receives the correct audio $X$, candidate list $C$, and reference transcript $Y^\star$, and generates only the caption:
\begin{equation}
Z_{\mathrm{off}}
\sim \pi_{\mathrm{SFT}}
\!\left(\cdot\mid X,C,Y^\star,\mathcal{D}_{3\mathrm{shot}}\right).
\end{equation}
We then assemble the generated caption with the reference transcription:
\begin{equation}
O^{\mathrm{off}}
=\left(
Z_{\mathrm{off}},
Y^\star
\right).
\end{equation}
The trajectory is off-policy because caption generation has access to the answer and its transcript is inserted from the reference. It is not sampled from the current policy under the original input $(X,C)$.

When injection is selected, we identify the rollout with the lowest combined reward in Eq.~\ref{eq:reward} and replace that single rollout with $O^{\mathrm{off}}$. We then recompute both reward components for all eight outputs. The injected output participates in the same reward normalization, advantage computation, and clipped loss implementation as the other seven outputs, without importance weighting for the behavior-policy mismatch. The resulting gradient is therefore biased relative to on-policy GRPO. We use the procedure as an empirical group-repair heuristic: it introduces one entity-complete trajectory while retaining seven sampled failures, whereas non-injected eligible groups preserve ordinary exploration.

The appendix gives the prompt construction and complete injection procedure.

\section{Experiments}

\begin{table*}[t]
\centering
\small
\setlength{\tabcolsep}{3.0pt}
\begin{tabular*}{\textwidth}{@{\extracolsep{\fill}}lcccc@{}}
\toprule
\multicolumn{5}{c}{\metriclegend} \\
\midrule
Model & Correct-only & Contextless & All-noise & RAG avg. \\
\midrule
\multicolumn{5}{l}{Models without explicit reasoning} \\
\addlinespace[2pt]
Qwen2-Audio & \trp{22.8}{19.0}{21.3} & \trp{23.3}{50.8}{59.5} & \trp{26.7}{51.4}{58.8} & \trp{28.9}{50.0}{34.9} \\
Qwen2.5-Omni-3B & \trp{4.2}{14.0}{14.8} & \trp{8.2}{45.2}{52.8} & \trp{8.8}{42.8}{50.5} & \trp{5.3}{22.2}{24.7} \\
Qwen2.5-Omni-7B & \trp{3.2}{11.4}{13.2} & \trp{7.2}{40.0}{46.4} & \trp{6.9}{37.1}{43.2} & \trp{4.4}{20.5}{22.6} \\
Qwen3-Omni-30B-Instruct & \trp{2.6}{9.2}{10.3} & \trp{\second{6.5}}{\second{38.0}}{\second{44.2}} & \trp{\second{6.5}}{\second{35.7}}{\second{42.2}} & \trp{3.6}{15.2}{17.2} \\
Step-Audio & \trp{4.4}{15.5}{16.1} & \trp{8.0}{45.7}{51.8} & \trp{9.7}{45.6}{51.5} & \trp{6.5}{19.2}{19.8} \\
Kimi-Audio & \trp{5.4}{25.2}{29.3} & \trp{7.2}{39.6}{45.9} & \trp{13.7}{41.5}{46.8} & \trp{10.3}{30.3}{30.5} \\
Instruct-SFT & \trp{\second{2.0}}{4.8}{4.8} & \trp{6.7}{39.7}{46.3} & \trp{6.6}{37.1}{44.0} & \trp{\second{3.4}}{14.9}{16.0} \\
Instruct-GRPO & \trp{\second{2.0}}{\second{4.5}}{\second{4.2}} & \trp{6.7}{39.8}{46.2} & \trp{\second{6.5}}{36.6}{43.2} & \trp{\second{3.4}}{\second{14.2}}{\second{15.8}} \\
\midrule
\multicolumn{5}{l}{Models with explicit reasoning} \\
\addlinespace[2pt]
Qwen3-Omni-30B-Think & \trp{3.3}{11.7}{13.1} & \trp{7.6}{43.4}{51.7} & \trp{8.4}{43.3}{51.2} & \trp{4.5}{18.1}{20.2} \\
Valley3 & \trp{4.7}{11.6}{12.5} & \trp{10.7}{52.3}{61.4} & \trp{11.8}{48.3}{56.6} & \trp{6.8}{20.7}{21.9} \\
Think-SFT & \trp{2.4}{12.5}{5.8} & \trp{7.4}{46.6}{53.1} & \trp{7.0}{43.0}{49.4} & \trp{3.8}{17.5}{18.5} \\
Think-GRPO & \trp{2.4}{5.4}{5.4} & \trp{7.5}{46.8}{53.4} & \trp{7.1}{46.3}{49.4} & \trp{3.8}{16.6}{18.2} \\
\CAST\ (ours) & \trpbest{1.9}{3.6}{3.4} & \trpbest{6.2}{34.6}{40.4} & \trpbest{6.3}{33.8}{40.0} & \trpbest{3.2}{11.8}{12.9} \\
\bottomrule
\end{tabular*}
\caption{Results on \RACE\ across external audio-language models and controlled 30B variants. Systems are grouped by whether they produce an explicit intermediate reasoning output. All metrics are percentages and lower is better. Bold and underlining mark the best and second-best values in each metric column. RAG is the arithmetic mean over top-$k\in\{5,10,20,50\}$.}
\label{tab:race_main}
\end{table*}

\subsection{Experimental Setup}
\label{sec:exp_setup}

We evaluate on \RACE\ and ContextASR-Bench \cite{ContextASR-Bench}. We report WER for overall transcription, named-entity word error rate (NE-WER) within aligned entity spans, and exact-match NE-FNR for missed ground-truth entities. Under irrelevant context, named-entity false-positive rate (NE-FPR) measures the proportion of output entities copied from the noise list. All values are percentages and lower is better; the appendix specifies normalization, alignment, denominator, and empty-context rules.

The baselines cover direct and reasoning-capable audio-language models at different scales: Qwen2-Audio \cite{qwen2audio}, Qwen2.5-Omni-3B and 7B \cite{qwen25omni}, Qwen3-Omni-30B-Instruct and Qwen3-Omni-30B-Think \cite{qwen3omni}, Step-Audio \cite{StepAudio}, Valley3 \cite{Valley3}, and Kimi-Audio \cite{KimiAudio}. Table~\ref{tab:race_main} includes every system evaluated on \RACE, while Table~\ref{tab:contextasr} includes systems with corresponding ContextASR-Bench results.

We construct controlled variants from the same 30B base model. Instruct-SFT and Instruct-GRPO transcribe directly without caption tags, whereas Think-SFT and Think-GRPO retain the base model's native reasoning format. \CAST-SFT uses the context-grounded caption without reinforcement learning. The no-injection baseline shares the \CAST-SFT checkpoint, caption format, 2,298 GRPO prompts, group size, reward, and optimization settings with \CAST, but uses ordinary rollout.

SFT uses 5,000 synthetic examples, and GRPO uses 298 post-SFT bad cases plus 2,000 new examples. The appendix reports the context-setting mixture, balanced retrieval sizes, and construction of these training records.

For SFT, we apply low-rank adaptation (LoRA) with rank 8 and scaling factor 32 to all linear layers while freezing the audio encoder and aligner. We train for three epochs with learning rate $5\times10^{-5}$, a 5\% warmup ratio, minimum learning rate $10^{-6}$, maximum sequence length 2,048, and global batch size 32. For GRPO, we retain the same LoRA configuration and train for three epochs with learning rate $10^{-5}$ and global batch size 128. We use the checkpoint at the end of this fixed training schedule rather than selecting one on an evaluation benchmark. We use $G=8$ generations per input, maximum input length 2,048, maximum completion length 1,024, temperature 1.0, top-$p$ 0.85, KL coefficient 0.001, and clipping parameter 0.2. Training uses eight GPUs with expert parallel size 8.

All supported models are deployed and evaluated with vLLM under matched settings. Models not supported by vLLM use the officially recommended inference implementation. Every context-capable system receives the same audio and candidate list, and evaluation uses the parsed transcript.

\subsection{Main Results}

Table~\ref{tab:race_main} first demonstrates why correct-only evaluation is insufficient. Qwen3-Omni-30B-Instruct reaches an NE-FNR of 10.3 when every target entity is supplied, but the value rises by 6.9 points under retrieved context and by 33.9 points without context. The correct-only score therefore substantially overestimates performance in a deployment-like retrieval setting. Entirely incorrect context exposes a different failure mode: Kimi-Audio's WER rises from 7.2 without context to 13.7 with an all-noise list. These two patterns are not interchangeable. Missing targets reveal dependence on oracle coverage, whereas the degradation from contextless to all-noise input reveals susceptibility to unsupported lexical cues.

\CAST\ achieves the lowest errors in every \RACE\ setting. The advantage is largest on the entity-sensitive metrics: its correct-only NE-FNR is 3.4, compared with 10.3 for Qwen3-Omni-30B-Instruct and 13.2 for Qwen2.5-Omni-7B. Under retrieved context, the corresponding values are 12.9, 17.2, and 22.6. This result is not obtained by indiscriminately following the list. \CAST\ also has the lowest contextless and all-noise WER, and the two values differ by only 0.1 point. The main comparison therefore supports both sides of selective contextualization: exploiting relevant evidence and remaining stable when no supplied candidate is correct.

\begin{table*}[t]
\centering
\small
\setlength{\tabcolsep}{3.0pt}
\begin{tabular*}{\textwidth}{@{\extracolsep{\fill}}lcccc@{}}
\toprule
\multicolumn{5}{c}{\metriclegend} \\
\midrule
Model & English Speech & English Dialogue & Mandarin Speech & Mandarin Dialogue \\
\midrule
\multicolumn{5}{l}{Models without explicit reasoning} \\
\addlinespace[2pt]
Qwen2-Audio & \trp{11.5}{27.3}{35.1} & \trp{14.0}{33.0}{32.9} & \trp{9.9}{24.1}{30.0} & \trp{7.0}{22.8}{26.2} \\
Qwen2.5-Omni-3B & \trp{4.0}{7.8}{9.7} & \trp{4.8}{14.4}{12.9} & \trp{2.1}{10.6}{14.1} & \trp{3.1}{15.1}{15.2} \\
Qwen2.5-Omni-7B & \trp{4.0}{7.4}{8.7} & \trp{5.3}{11.8}{9.2} & \trp{1.8}{9.8}{12.2} & \trp{2.4}{14.1}{13.2} \\
Kimi-Audio & \trp{2.9}{6.7}{8.0} & \trp{4.7}{13.5}{11.3} & \trp{2.0}{11.1}{15.3} & \trp{2.9}{15.9}{16.7} \\
Qwen3-Omni-30B-Instruct & \trp{\best{1.1}}{1.7}{0.3} & \trp{4.4}{21.3}{1.2} & \trp{0.8}{5.4}{0.4} & \trp{1.6}{13.4}{0.7} \\
Instruct-SFT & \trp{\best{1.1}}{\second{1.2}}{\best{0.0}} & \trp{2.0}{\second{4.0}}{\second{0.3}} & \trp{\best{0.4}}{\second{4.3}}{\best{0.1}} & \trp{\best{0.7}}{7.6}{\second{0.3}} \\
Instruct-GRPO & \trp{\best{1.1}}{\second{1.2}}{\second{0.1}} & \trp{\second{1.9}}{\best{3.9}}{\best{0.2}} & \trp{\best{0.4}}{\second{4.3}}{\best{0.1}} & \trp{\best{0.7}}{7.6}{\best{0.2}} \\
\midrule
\multicolumn{5}{l}{Models with explicit reasoning} \\
\addlinespace[2pt]
Think-SFT & \trp{1.4}{\second{1.2}}{\second{0.1}} & \trp{2.4}{4.7}{\second{0.3}} & \trp{\second{0.5}}{\best{4.2}}{\second{0.2}} & \trp{\second{0.8}}{\second{7.5}}{\second{0.3}} \\
Think-GRPO & \trp{\second{1.3}}{\second{1.2}}{\second{0.1}} & \trp{2.4}{5.2}{\second{0.3}} & \trp{\second{0.5}}{\second{4.3}}{\best{0.1}} & \trp{\second{0.8}}{7.6}{\second{0.3}} \\
\CAST\ (ours) & \trp{\best{1.1}}{\best{1.1}}{\second{0.1}} & \trp{\best{1.8}}{4.4}{\best{0.2}} & \trp{\second{0.5}}{4.7}{\best{0.1}} & \trp{\second{0.8}}{\best{7.0}}{\best{0.2}} \\
\bottomrule
\end{tabular*}
\caption{Results on ContextASR-Bench for systems with corresponding evaluations, grouped by whether they produce an explicit intermediate reasoning output. All metrics are percentages and lower is better; bold and underlining mark the best and second-best values in each metric column.}
\label{tab:contextasr}
\end{table*}

Table~\ref{tab:contextasr} reports the corresponding ContextASR-Bench results. \CAST\ obtains NE-FNR values of 0.1, 0.2, 0.1, and 0.2 across its four subsets, improving over Qwen3-Omni-30B-Instruct in each case. On English Dialogue, it reduces NE-WER from 21.3 to 4.4 and WER from 4.4 to 1.8. Together with Table~\ref{tab:race_main}, these results show strong recognition with correct context as well as robustness when context is missing, retrieved, or incorrect.

\subsection{Retrieval-Augmented Context}

\begin{figure}[!t]
\centering
\includegraphics[width=\columnwidth]{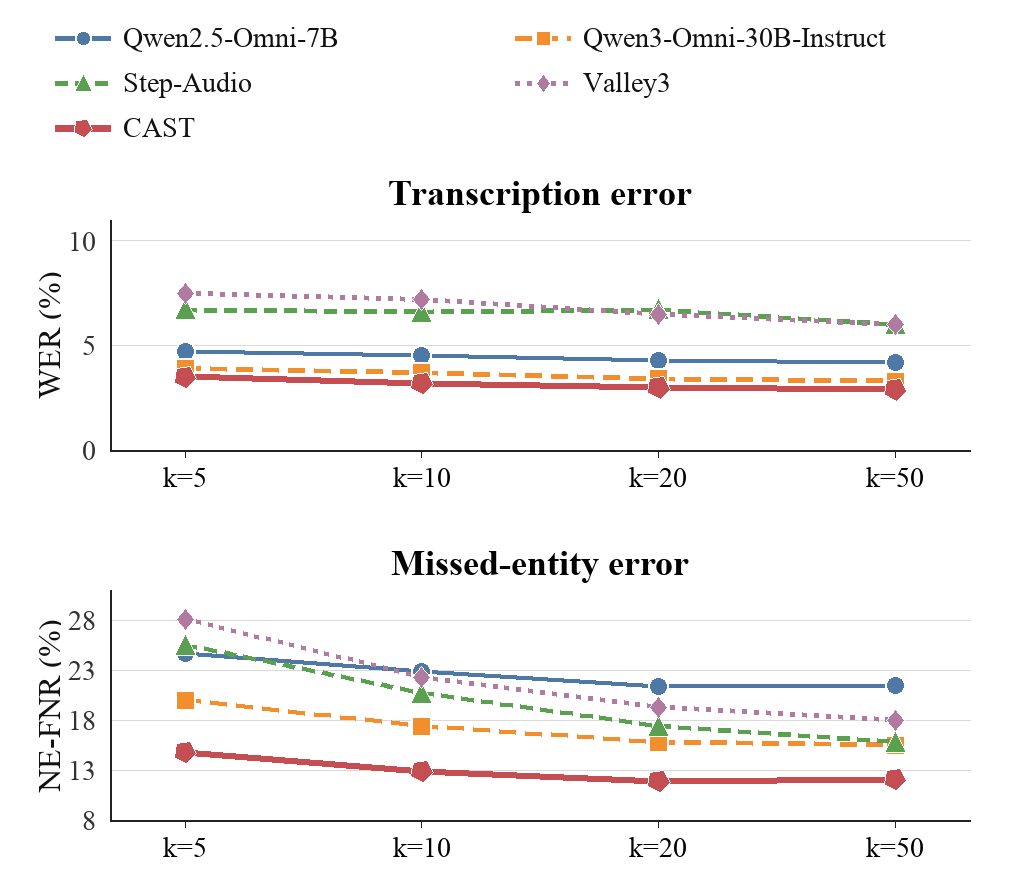}
\caption{WER and NE-FNR as retrieval size increases on \RACE.}
\label{fig:rag_k}
\end{figure}

Increasing $k$ improves the probability that a target entity is retrieved, but it also enlarges the set of distractors that may bias decoding. Figure~\ref{fig:rag_k} shows that \CAST\ remains strongest in both WER and NE-FNR across all retrieval sizes. Its transcription error stays stable while its missed-entity error decreases as retrieval coverage grows, followed by only mild saturation at the largest list. Qwen3-Omni-30B-Instruct also benefits from larger lists, but retains a consistent gap, while Valley3 and Step-Audio show weaker or less regular improvements. \CAST\ therefore converts broader hotword coverage into recognition gains without allowing the additional candidates to dominate the acoustic evidence.

\subsection{Incorrect-Context Robustness}

\begin{figure}[t]
\centering
\includegraphics[width=\columnwidth]{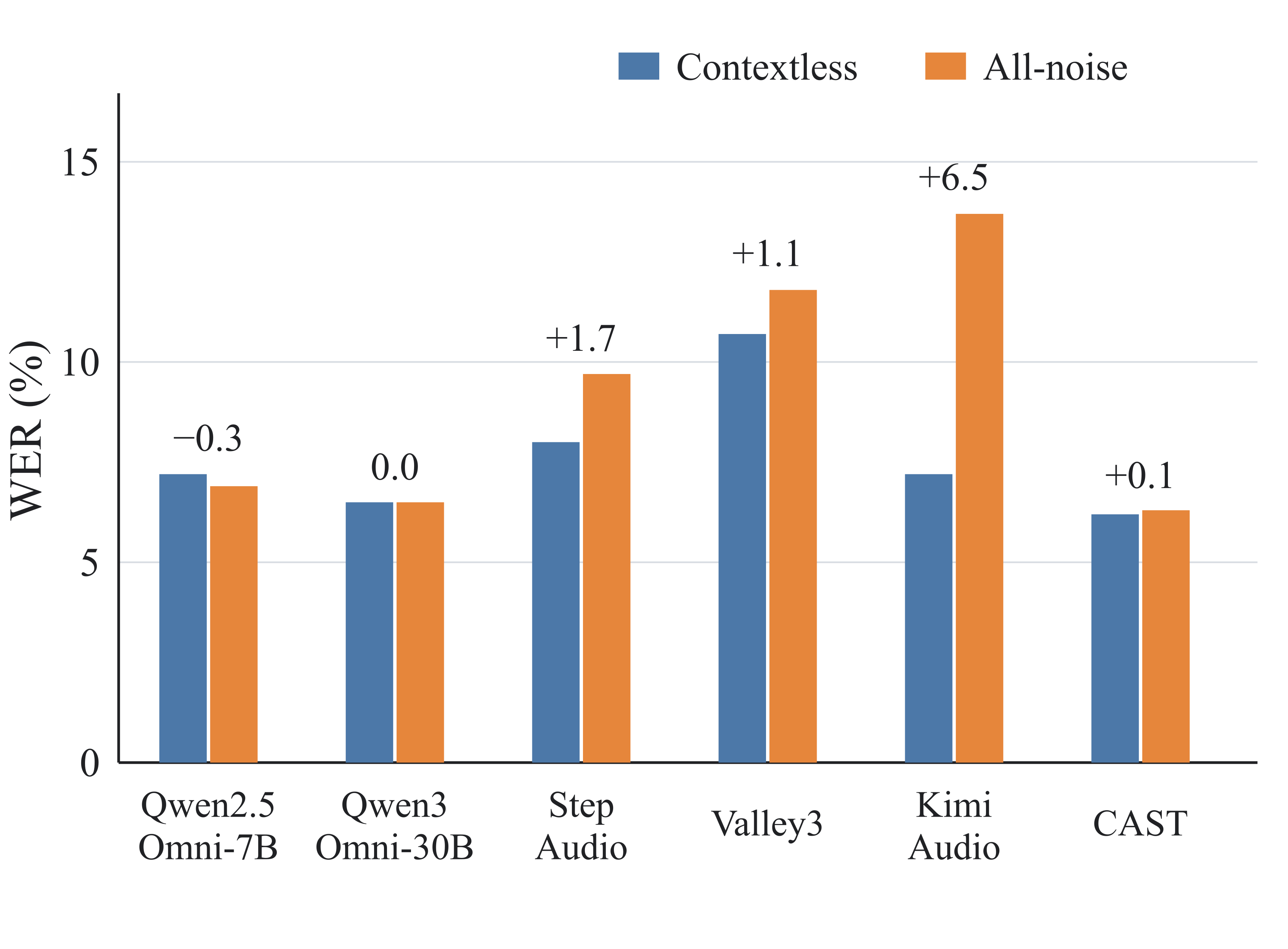}
\caption{WER under contextless and all-noise input on \RACE.}
\label{fig:noise_robustness}
\end{figure}

Figure~\ref{fig:noise_robustness} tests whether an entirely incorrect list distorts the transcript. \CAST\ changes minimally from contextless to all-noise input, whereas Kimi-Audio, Step-Audio, and Valley3 exhibit clear WER degradation. NE-FPR provides complementary evidence by measuring whether entities are copied directly from the noise list. Although Qwen2.5-Omni-7B attains a slightly lower NE-FPR, it has weaker entity recall when relevant context is available. \CAST\ combines a low false-activation rate with the strongest all-noise NE-FNR among the displayed systems, indicating that its robustness does not come from ignoring contextual information altogether.

\subsection{Ablations}

Table~\ref{tab:inject_comparison} isolates trajectory injection against the matched GRPO baseline.

\begin{table}[t]
\centering
\small
\setlength{\tabcolsep}{2.0pt}
\begin{tabular}{@{}lccc@{}}
\toprule
\multicolumn{4}{c}{\metriclegend} \\
\midrule
Setting & Inject & Baseline & $\Delta$ \\
\midrule
Correct-only & \trpbest{1.9}{3.6}{3.4} & \trp{2.5}{7.6}{8.3} & \trp{-0.6}{-4.0}{-4.9} \\
Contextless & \trpbest{6.2}{34.6}{40.4} & \trp{6.9}{39.8}{46.3} & \trp{-0.7}{-5.2}{-6.0} \\
All-noise & \trpbest{6.3}{33.8}{40.0} & \trp{6.8}{36.9}{43.7} & \trp{-0.5}{-3.2}{-3.7} \\
RAG avg. & \trpbest{3.2}{11.8}{12.9} & \trp{3.7}{14.7}{16.5} & \trp{-0.5}{-2.9}{-3.6} \\
\bottomrule
\end{tabular}
\caption{Direct comparison between trajectory injection and the otherwise matched GRPO baseline on \RACE. All metrics are percentages and lower is better. Bold marks the better system; $\Delta$ is inject minus baseline, so negative values favor injection.}
\label{tab:inject_comparison}
\end{table}

Injection improves all three metrics in every setting relative to the otherwise matched GRPO baseline. The improvements extend to contextless and all-noise evaluation, even though these inputs contain no supplied target entity and cannot trigger replacement during training. This suggests that repairing eligible all-failure groups changes the learned transcription policy rather than merely memorizing the injected examples. The complete ablations in the appendix separate direct transcription, caption supervision, GRPO, injection, and the two reward terms. Together, they show that the caption defines the intended intermediate behavior, while targeted injection is needed to turn that behavior into consistent entity-recognition gains without sacrificing transcript fidelity.

\subsection{Discussion}

The main results distinguish task-specific contextual reasoning from unrestricted chain-of-thought. The native Qwen3-Omni-30B-Think model and Valley3 both generate explicit reasoning, yet remain behind \CAST\ on contextual entity recognition. Think-SFT and Think-GRPO use the same 30B base model and task data but still do not match \CAST, showing that additional training of a generic reasoning format is insufficient. The Instruct-SFT and Instruct-GRPO variants are more competitive in several correct-context conditions, but their \RACE\ results remain weaker when retrieval is incomplete or incorrect. The ContextASR-Bench results further show that the advantage of the context-grounded caption is not confined to the newly constructed benchmark. These comparisons concern transcription behavior rather than caption faithfulness, since the caption receives no independent correctness reward.

\subsubsection{Output Length and Estimated Latency}

\begin{table}[t]
\centering
\small
\setlength{\tabcolsep}{3.5pt}
\begin{tabular}{@{}lrr@{}}
\toprule
Model & Avg. tokens & Est. decode (s) \\
\midrule
Ground-truth transcript & 27.3 & -- \\
Qwen2.5-Omni-3B & 27.8 & 0.2 \\
Qwen2.5-Omni-7B & 27.7 & 0.3 \\
Qwen3-Omni-30B-Instruct & 27.9 & 0.2 \\
\midrule
Qwen3-Omni-30B-Think & 892.4 & 6.5 \\
Valley3 & 245.7 & 3.0 \\
\midrule
\CAST\ (ours) & 65.1 & 0.5 \\
\bottomrule
\end{tabular}
\caption{Average output length and approximate decoding cost.}
\label{tab:output_cost}
\end{table}

\CAST\ averages 65.1 output tokens, compared with 892.4 for Qwen3-Omni-30B-Think and 245.7 for Valley3. To estimate decoding cost, we use the official batch-one BF16 speeds of Qwen3-4B, Qwen3-8B, and Qwen3-30B-A3B on one NVIDIA H20 as proxies for the 3B, 7--8B, and 30B-A3B systems, respectively \cite{QwenSpeedBenchmark}; the estimates exclude audio encoding and prefill. \CAST\ is longer than direct transcription because it includes a caption, but remains much closer to the reference-transcript length than the unrestricted reasoning models. Its estimated decoding cost is also substantially lower, giving it a stronger accuracy--efficiency trade-off. These values are analytical estimates rather than end-to-end measurements, and actual latency depends on the serving stack, batching, and audio length.

\section{Limitations}

\RACE\ currently covers only English and Chinese, so broader linguistic and acoustic variation remains untested. In addition, \CAST\ is evaluated with a 30B-A3B base model; whether the same reasoning and injection strategy remains effective for substantially smaller models requires further study.

\section{Conclusion}

We introduce \RACE, a 9,136-utterance benchmark that evaluates contextual ASR under correct-only, contextless, retrieved, and entirely incorrect candidate lists. We propose \CAST, whose context-grounded caption provides a compact reasoning step for using supported hotwords while rejecting unsupported ones. We further introduce targeted off-policy trajectory injection to repair GRPO groups in which every rollout misses a supplied entity. Experiments on \RACE\ and ContextASR-Bench show that \CAST\ improves contextual entity recognition and retrieval robustness while remaining substantially more efficient than unrestricted audio reasoning.

\bibliography{references}

\end{document}